\documentclass[conference, a4paper]{IEEEtran}
\usepackage{cite}
\usepackage{amsmath,amssymb,amsfonts}
\usepackage{graphicx}
\usepackage{textcomp}
\usepackage{booktabs}
\usepackage{multirow}
\usepackage{array}
\usepackage[T1]{fontenc}
\usepackage[colorlinks=true,linkcolor=black,citecolor=black,urlcolor=blue]{hyperref}
\graphicspath{{./}}

\begin{document}

\title{Interpretable Factor Decomposition for Decision Intelligence in Large-Scale Financial Markets: Evidence from China's A-Share Market}

\author{\IEEEauthorblockN{1\textsuperscript{st} Xiao Han}
\IEEEauthorblockA{\textit{Goizueta Business School} \\
\textit{Emory University}\\
Atlanta, GA, USA \\
xhan@alumni.emory.edu}
\and
\IEEEauthorblockN{2\textsuperscript{nd} Yao Xiao}
\IEEEauthorblockA{\textit{College of Computing} \\
\textit{Georgia Institute of Technology}\\
Atlanta, GA, USA \\
yxiao344@gatech.edu}
\and
\IEEEauthorblockN{3\textsuperscript{rd} Zhen Zhang}
\IEEEauthorblockA{\textit{School of Data Science} \\
\textit{University of Pennsylvania}\\
Philadelphia, PA, USA \\
billzhangzhen98@gmail.com}
\and
\IEEEauthorblockN{4\textsuperscript{th} Moxuan Zheng}
\IEEEauthorblockA{\textit{Stern School of Business} \\
\textit{New York University}\\
New York, NY, USA \\
mz2156@nyu.edu}
}

\maketitle

\begin{abstract}
We present an interpretable machine learning pipeline to decompose cross-sectional equity return predictability into auditable factor contributions. We apply an XGBoost model with TreeSHAP attribution and conduct stress testing on 3,632 Chinese A-share stocks from 2009 until 2019. On prediction, using 60-month rolling windows over 55 months of out-of-sample data, XGBoost obtains a mean AUC of 0.547 (rank IC = 0.119) and $+2.38\%$/month (Newey-West $t = 5.94$; annualized Sharpe 2.23) long-short spread for the top vs bottom quintiles. This alpha is persistent after adjusting for the Carhart four-factor model ($+2.31\%$/month; $t = 7.48$). On interpretation, SHAP decomposition indicates that behavioral signals (turnover and momentum) account for 58.2\% of predictive attribution compared to 10.7\% for valuation ratios, on average, across 50 industry groups. Ablation analysis serves to cross-validate this ranking and provides evidence that SHAP and ablation diverge in a manner that highlights feature substitutability structure that is largely invisible to either method used in isolation.
\end{abstract}


\section{Introduction}

Which financial indicators predict cross-sectional stock returns, and how much does each contribute when modeled jointly? Starting with value investing analysis---testing the outperformance of low P/E, P/B, and P/CF stocks---we confirmed that low-valuation stocks outperformed; however, we found that behavioral signals (turnover and momentum) consistently ranked higher than valuation ratios when assessed together. We describe this process as a \textit{decision intelligence} analysis where we take raw financial indicators and transform them into transparent and auditable decision signals. TreeSHAP~\cite{lundberg2020} provides exact and additive attribution satisfying basic fairness criteria to ensure that the decomposition is accurate and not only heuristic. We further complement this work via an ablation analysis, which acts as a standalone basis of assessment while adding value as a diagnostic.

We assess via the Chinese A-share market of over 3,600 firms trading basically 80\%~\cite{ng2007} by retailers to test the robustness of the attribution method by its ability to adaptively stress test from significant behavioral signals and multiple regime shifts (2009--2019).

There are three major findings. First, SHAP assigns 58.2\% of predictive attribution to behavioral indicators (turnover, momentum, volatility) and only 10.7\% to valuation ratios, a pattern that holds across all calendar years and 50 industry groups with sufficient depth for portfolio construction. Second, XGBoost outperforms both logistic regression and single-factor sorts, producing an out-of-sample AUC of 0.547, a long-short spread of $+2.38\%$/month (Carhart four-factor alpha $+2.31\%$/month, $t = 7.48$), and an annualized Sharpe ratio of 2.23. Third, while both SHAP and ablation provide relevant feature-level diagnostics, side-by-side comparison reveals that certain features are load-bearing (e.g., size) while others are substitutable (e.g., turnover), with a Spearman rank correlation of $\rho = 0.54$ between the two importance orderings. This distinction between preferred and necessary features has direct implications for production decision system design.

\section{Related Work}

\textbf{Factor investing and anomalies.} The literature stretches back to the Capital Asset Pricing Model (CAPM) through the work of Fama and French~\cite{fama1993} to what Harvey, Liu, and Zhu~\cite{harvey2016} have described as the ``zoo'' of over 300 documented anomalies. Carhart~\cite{carhart1997} added the momentum factor and Fama and French~\cite{fama2015} extended to a five-factor model; Jegadeesh and Titman~\cite{jegadeesh1993} developed the momentum anomaly, and Asness, Moskowitz, and Pedersen~\cite{asness2013} showed the generality of cross-asset-class momentum.

\textbf{Machine learning in asset pricing.} Gradient boosting, introduced by Friedman~\cite{friedman2001}, forms the basis for most recent ML applications in asset pricing. Gu, Kelly, and Xiu~\cite{gu2020} benchmarked ML methods for empirical asset pricing and showed that tree-based models outperform linear models significantly, with typical out-of-sample AUC of 0.53--0.56, establishing the relevant performance ceiling for our results. Deep temporal architectures, including TCN-based frameworks with walk-forward evaluation, have also been applied to equity forecasting~\cite{yang2025}; however, our cross-sectional ranking task favors tree ensembles for their native handling of heterogeneous tabular features, and we use XGBoost~\cite{chen2016} as our primary model.

\textbf{Explainable AI and SHAP.} SHAP~\cite{lundberg2017} provides a game-theoretic foundation for attributing features; TreeSHAP~\cite{lundberg2020} computes exact Shapley values for tree ensembles in polynomial time. Rudin~\cite{rudin2019} argues that inherently interpretable models are preferable in high-stakes situations; we adopt her interpretability motivation while using TreeSHAP, which provides exact (not approximate) attribution for tree-based models.

\textbf{China A-share return predictability.} Liu, Stambaugh, and Yuan~\cite{liu2019} developed the Chinese three-factor model, Carpenter, Lu, and Whitelaw~\cite{carpenter2021} found that the market is useful in forecasting future profits, and Chen, Chao, and Wu~\cite{chen2021} showed that turnover has a negative effect on subsequent returns.

No prior work, to our knowledge, systematically quantifies the SHAP-ablation rank divergence as a diagnostic for feature dependency structure in financial prediction.

\section{Data and Methodology}

\subsection{Data Sources and Panel Construction}

Utilizing the baostock API (database), we built a new daily panel consisting of firm-level observations on all CSI All-Share Index stocks. After excluding ST-flagged securities, the total stock universe from 2009--2019 contained 3,632 individual securities. We then created a monthly panel by aggregating the daily data. The monthly panel includes 13 columns corresponding to the following categories: (1)~\textit{Valuation}: Mean P/E, P/B, P/S, P/CF; (2)~\textit{Behavioral}: Monthly return, mean turnover, intra-month volatility, 3-/6-/12-month momentum; (3)~\textit{Fundamental}: ROE, NP margin (quarterly, forward-filled with a one-quarter lag to approximate point-in-time availability, 88.4\% covered); (4)~\textit{Size}: Log of Mkt Cap. Excluded stock/months with $<$10 trading days from the monthly summary. The 1st and 99th percentiles for each feature in the monthly summary were winsorized (acceptable data cleaned to remove extreme values). In the end, we had a panel with 254,854 stock/month observations (3,199 stocks, 116 months).

\subsection{Prediction Task and Model}

The goal is to find out if a stock returns more than the median return of all stocks in a one-month period using a binary measure in the model for the rankings.

The training of the model via XGBoost~\cite{chen2016} is accomplished using fixed hyperparameters across all windows (200 trees, depth 4, learning rate 0.05, and 80 percent row and column subsampling) to eliminate implicit look-ahead bias in data-driven parameter selection, following Gu, Kelly, and Xiu~\cite{gu2020}. Sensitivity tests with alternative configurations (depth 3/300 trees; depth 5/150 trees; learning rate 0.10) yield mean AUC in the range 0.544 to 0.549, within the reported confidence interval, confirming that results are not sensitive to the specific hyperparameter choice. Validation is performed using rolling windows---60 months for training, then one month for testing---following strict temporal separation, which recent work has identified as a common source of inflated performance claims when not enforced~\cite{yao2026}. We report bootstrap confidence intervals rather than single-point estimates, given evidence that single-seed evaluation can yield unreliable metric estimates in predictive modeling~\cite{zhan2026}.

\subsection{SHAP Attribution and Ablation}

TreeSHAP~\cite{lundberg2020} decomposes each out-of-sample prediction into exact, additive feature contributions. For each rolling window, SHAP values are computed on the test-month predictions (not on training data), and global importance is the mean absolute SHAP value aggregated across all 55 out-of-sample months. The validation of the model for independent validation is performed using the process of ablations, which is defined as the systematic elimination of feature groups from the model and re-running the entire statistical model using the walk-forward validation process, with the goal of measuring both the change in AUC and the difference between the predictive model rankings (SHAP values) and rankings from the ablation process.

\subsection{Portfolio Construction and Baselines}

The evaluation of the predicted probabilities used to create quintile portfolios consists of evaluating the portfolios as equal-weighted, float-cap-weighted, and industry-neutral via ranking within their respective CSRC industry group. The transaction costs incurred to create each quintile portfolio round-trip were as follows: [0, 0.2\%, 0.6\%, 1.0\%] percent. The baseline evaluation for the five portfolios consists of equal-weighted portfolios among all stocks, single-factor sorts of all stocks, and a logistic regression model of the return using the thirteen features used for prediction. Because short-selling is restricted in China's A-share market, the long-short spread is reported as a measure of cross-sectional signal strength; we also report the long-only top quintile excess return over the equal-weight benchmark.

\subsection{Implementation}

The pipeline is implemented in Python 3.8 using XGBoost 1.7, SHAP 0.41 for TreeSHAP computation, and statsmodels 0.13 for Newey-West and factor-model regressions. All code and reproduction instructions are publicly available.\footnote{https://github.com/XH8851/Interpretable-Factor-Decomposition-for-Decision-Intelligence-in-Large-Scale-Financial-Markets}

Descriptive statistics for the panel are provided in Table~\ref{tab:descriptive}.

\begin{table}[htbp]
\caption{Descriptive Statistics ($N = 254{,}854$ stock-months)}
\begin{center}
\footnotesize
\setlength{\tabcolsep}{3pt}
\begin{tabular}{lccccc}
\toprule
\textbf{Variable} & \textbf{Obs} & \textbf{Mean} & \textbf{Std.} & \textbf{Min} & \textbf{Max} \\
\midrule
Monthly return (\%) & 254,854 & 0.76 & 13.36 & $-$47.98 & 141.12 \\
Mean P/E & 254,854 & 64.37 & 191.09 & $-$2,391 & 2,931 \\
Mean P/B & 254,854 & 4.51 & 4.27 & 0.67 & 64.03 \\
Mean turnover (\%) & 254,854 & 2.83 & 2.95 & 0.08 & 24.99 \\
Volatility (\%) & 254,854 & 2.62 & 1.31 & 0.00 & 8.73 \\
12-mo momentum (\%) & 254,854 & 12.14 & 46.95 & $-$112.69 & 352.44 \\
ROE (avg) & 254,854 & 0.05 & 0.08 & $-$1.15 & 0.44 \\
Net profit margin & 254,854 & 0.08 & 0.17 & $-$2.69 & 0.70 \\
Log market cap & 254,854 & 31.75 & 0.97 & 29.63 & 35.44 \\
\bottomrule
\multicolumn{6}{l}{\scriptsize Note: Features winsorized at 1st/99th percentiles per monthly cross-section.} \\
\end{tabular}
\label{tab:descriptive}
\end{center}
\end{table}

\section{Results}

\subsection{ML Performance and Model Comparison}

Table~\ref{tab:cv} shows the results each year. XGBoost produced positive long-short returns and AUCs in each of the five years, with 2015 being the weakest year and 2018 the strongest. Over the full 55-month period, 80\% of months are profitable (hit rate), maximum drawdown is 7.8\%, and the long-only top quintile outperforms the equal-weight benchmark by $+1.48\%$/month, confirming signal value without short positions. The mean monthly rank IC (Spearman correlation between predicted probability and realized return) is 0.119 with an information coefficient ratio (ICIR) of 1.12 ($t = 8.26$), and 56.4\% of stocks placed in the top quintile by the model actually outperform the cross-sectional median (top-quintile precision). Monthly one-way portfolio turnover in the top quintile averages 58.3\%, which at a 30 bps one-way cost implies an effective round-trip drag of approximately 0.35\%/month, consistent with the transaction cost analysis in Section~IV-F. Although an AUC of 0.547 appears modest by conventional classification standards, cross-sectional return prediction is a low-signal environment with inherently noisy labels~\cite{li2025theory} where market efficiency caps AUC well below 0.60~\cite{gu2020}. The portfolio exploits the tails of the predicted distribution (top and bottom quintiles), where small ranking improvements translate into economically large return differentials.

\begin{table}[htbp]
\caption{Monthly Rolling Cross-Validation Results (by Year)}
\begin{center}
\footnotesize
\begin{tabular}{lccccc}
\toprule
\textbf{Year} & \textbf{$N$ mo} & \textbf{AUC} & \textbf{L-S (\%/mo)} & \textbf{Std (\%)} & \textbf{Sharpe} \\
\midrule
2015 & 12 & 0.516 & $+$0.81 & 5.06 & 0.55 \\
2016 & 12 & 0.549 & $+$2.95 & 3.91 & 2.61 \\
2017 & 12 & 0.543 & $+$2.53 & 1.82 & 4.83 \\
2018 & 12 & 0.568 & $+$3.09 & 2.40 & 4.47 \\
2019 & 7  & 0.558 & $+$1.87 & 4.92 & 1.32 \\
\midrule
\textbf{Overall} & \textbf{55} & \textbf{0.547} & \textbf{$+$2.38} & \textbf{3.67} & \textbf{2.23} \\
\bottomrule
\multicolumn{6}{l}{\scriptsize 95\% bootstrap CIs: AUC [0.532, 0.563]; L-S [$+$1.44\%, $+$3.34\%];} \\
\multicolumn{6}{l}{\scriptsize Sharpe [1.38, 3.36]. Newey-West $t = 5.94$ ($p < 0.001$).} \\
\end{tabular}
\label{tab:cv}
\end{center}
\end{table}

Table~\ref{tab:model_comp} compares models and construction methods. XGBoost provided a return in AUC 0.46\% higher than LR, and industry neutral sorting of LR improved the Sharpe for LR by 0.64 versus an equal-weighted LR.

\begin{table}[htbp]
\caption{Model and Construction Scheme Comparison (55 OOS months)}
\begin{center}
\footnotesize
\begin{tabular}{llcc}
\toprule
\textbf{Panel} & \textbf{Configuration} & \textbf{L-S (\%/mo)} & \textbf{Sharpe} \\
\midrule
\multicolumn{4}{l}{\textit{A. Model comparison (equal-weight construction)}} \\
& Equal-weight market & $+$0.90 & -- \\
& Best single factor (turnover) & $+$0.73 & 0.61 \\
& Logistic Regression & $+$1.92 & 1.34 \\
& \textbf{XGBoost (ours)} & \textbf{$+$2.38} & \textbf{2.23} \\
\midrule
\multicolumn{4}{l}{\textit{B. Construction scheme (LR ranking, AUC\,=\,0.546)}} \\
& Equal-weight (reference) & $+$1.92 & 1.34 \\
& Float-cap-weighted & $+$2.21 & 1.10 \\
& Industry-neutral & $+$2.12 & 1.98 \\
\bottomrule
\end{tabular}
\label{tab:model_comp}
\end{center}
\end{table}

\begin{figure}[htbp]
\centerline{\includegraphics[width=\columnwidth]{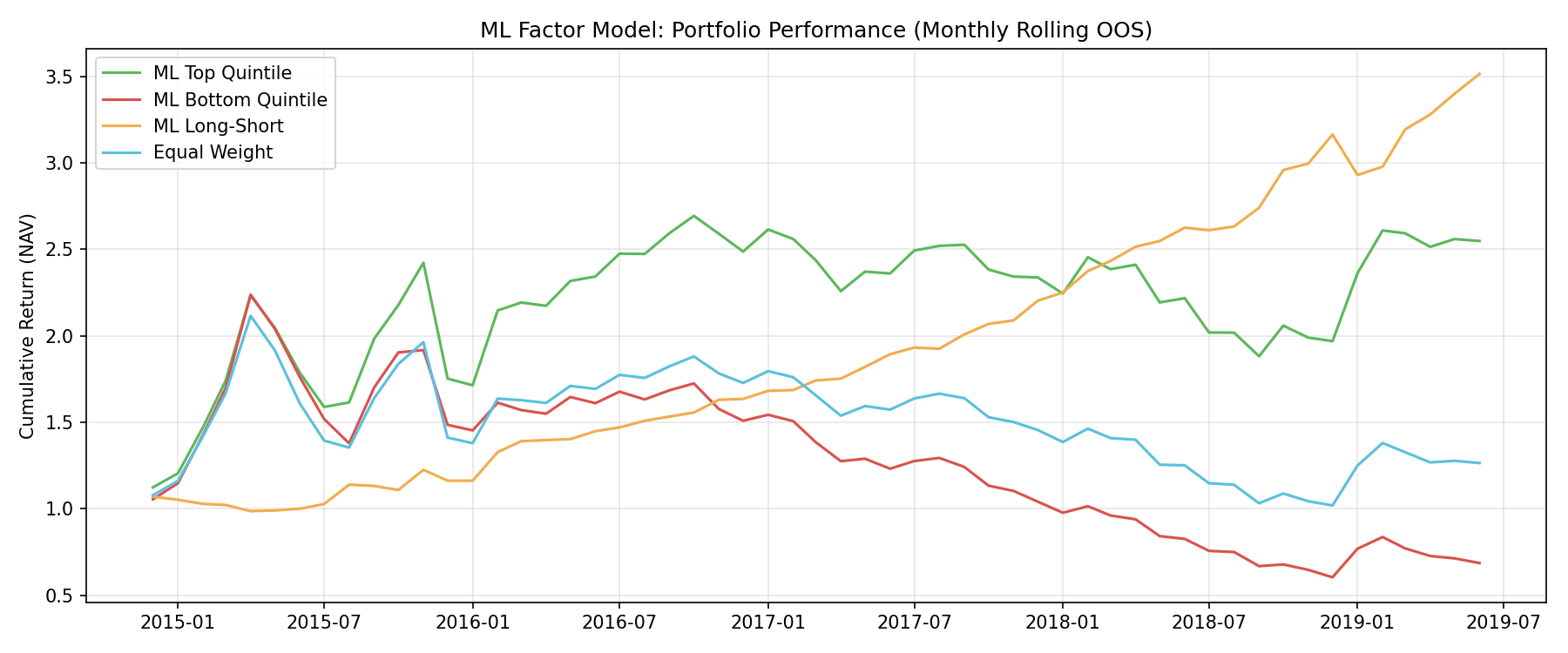}}
\caption{Cumulative portfolio returns: ML top/bottom quintile, long-short, and equal-weight benchmark over 55 out-of-sample months (Dec 2014 to Jun 2019).}
\label{fig:portfolio}
\end{figure}

\subsection{Factor-Model Alpha}

Table~\ref{tab:alpha} provides regression results using factor-models. The CAPM beta is approximately 0.0 ($\pm 0.03$, $t = -0.27$) while the Carhart returns alpha is 2.31\% per month ($t = 7.48$).

\begin{table}[htbp]
\caption{Factor-Model Alpha (HAC Standard Errors)}
\begin{center}
\footnotesize
\begin{tabular}{lcccc}
\toprule
\textbf{Model} & \textbf{Alpha (\%/mo)} & \textbf{$t$-stat} & \textbf{$p$-value} & \textbf{$R^2$} \\
\midrule
CAPM & $+$2.40 & 7.54 & $<$0.001 & 0.001 \\
Fama-French 3 & $+$2.31 & 7.65 & $<$0.001 & 0.005 \\
Carhart 4 & $+$2.31 & 7.48 & $<$0.001 & 0.021 \\
\bottomrule
\end{tabular}
\label{tab:alpha}
\end{center}
\end{table}

\subsection{SHAP Feature Decomposition}

\begin{figure}[htbp]
\centerline{\includegraphics[width=\columnwidth]{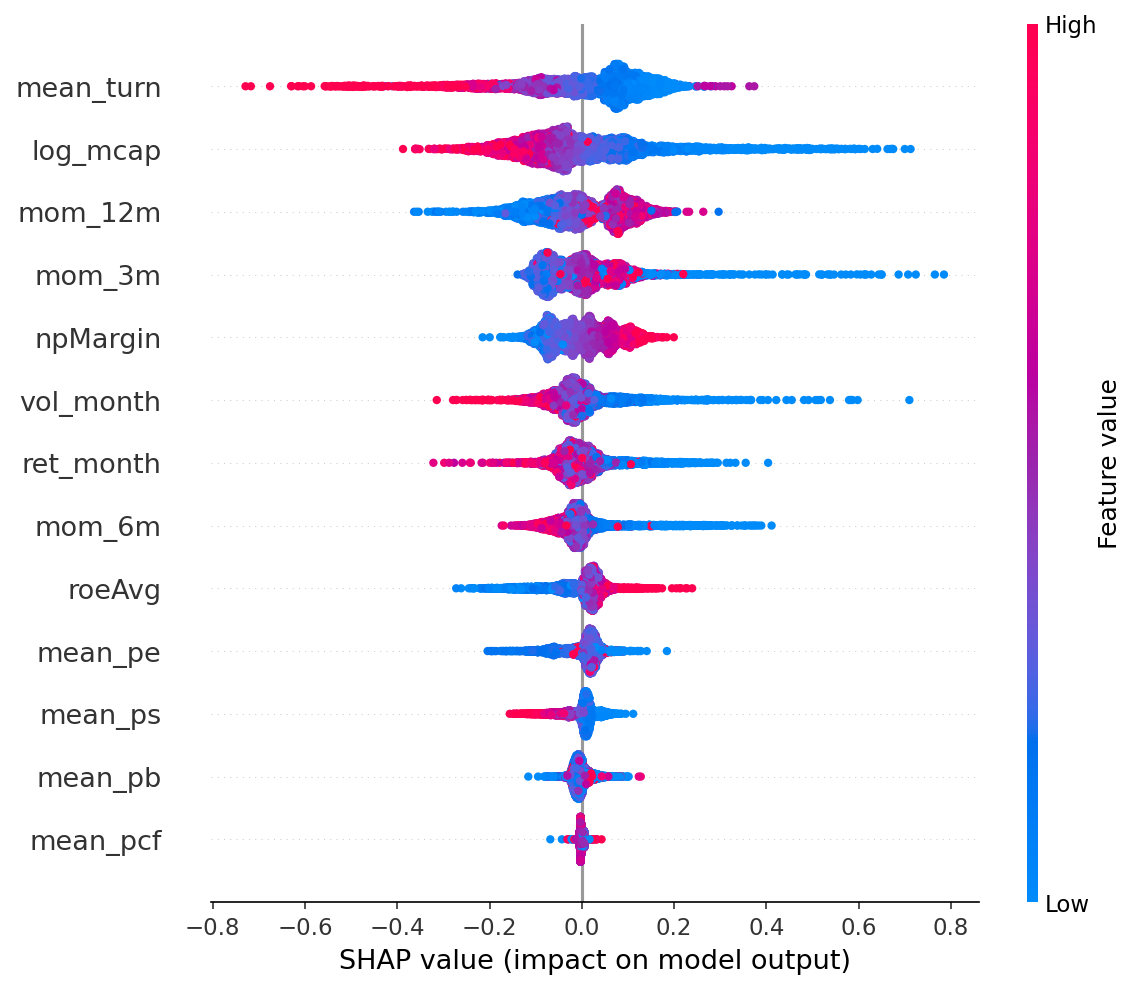}}
\caption{SHAP summary (beeswarm) plot showing feature importance and directional effects for all 13 features.}
\label{fig:shap}
\end{figure}

Fig.~\ref{fig:shap} displays the SHAP beeswarm plot for all 13 features. Each dot represents one stock-month observation; horizontal position indicates the signed SHAP contribution, and color encodes the raw feature value (red = high, blue = low). The plot reveals that high turnover pushes predictions toward outperformance (red dots shift right), while high log market cap pushes predictions toward underperformance (red dots shift left), consistent with the well-documented size effect. Momentum features show a nonlinear pattern: extreme positive momentum contributes positively, but moderate values cluster near zero.

Table~\ref{tab:shap} calculates global SHAP contribution; behavioral features contribute to 58.2\% of total attribution, size features contributed 16.4\%, fundamental features contributed 14.7\%, while value features contributed to 10.7\%. A major part of the turnover feature aligns with the behavioral finance literature which points to the importance of attention~\cite{barber2008}, liquidity, and the relationship between illiquidity and returns~\cite{datar1998,amihud2002,brennan1998,lee2000}.

\begin{table}[htbp]
\caption{SHAP Feature Importance (Global)}
\begin{center}
\footnotesize
\setlength{\tabcolsep}{3pt}
\begin{tabular}{clccc}
\toprule
\textbf{Rank} & \textbf{Feature} & \textbf{$|\text{SHAP}|$} & \textbf{Category} & \textbf{\%} \\
\midrule
1 & Mean turnover & 0.1091 & Behavioral & 16.5 \\
2 & Log market cap & 0.1086 & Size & 16.4 \\
3 & 12-mo momentum & 0.0753 & Behavioral & 11.4 \\
4 & 3-mo momentum & 0.0622 & Behavioral & 9.4 \\
5 & Net profit margin & 0.0558 & Fundamental & 8.4 \\
6 & Monthly volatility & 0.0519 & Behavioral & 7.8 \\
7 & Monthly return & 0.0432 & Behavioral & 6.5 \\
8 & 6-mo momentum & 0.0426 & Behavioral & 6.4 \\
9 & ROE (avg) & 0.0411 & Fundamental & 6.2 \\
10 & Mean P/E & 0.0313 & Valuation & 4.7 \\
11 & Mean P/S & 0.0228 & Valuation & 3.4 \\
12 & Mean P/B & 0.0133 & Valuation & 2.0 \\
13 & Mean P/CF & 0.0036 & Valuation & 0.5 \\
\bottomrule
\multicolumn{5}{l}{\scriptsize Percentages rounded individually; sums may differ from aggregates due to rounding.} \\
\end{tabular}
\label{tab:shap}
\end{center}
\end{table}

\subsection{Ablation Study}

Table~\ref{tab:ablation} provides ablative results for each of the features used in this study. Log market cap had the greatest discriminative power with the AUC dropping to $-0.009$ when log market cap was left out. Removal of metrics that reflect value from consideration has a minor positive impact on AUC (e.g., $+0.001$), since these types of features obtain low variable coverage and therefore add noise to the AUC findings.

\begin{table}[htbp]
\caption{Ablation Results}
\begin{center}
\footnotesize
\begin{tabular}{lcc}
\toprule
\textbf{Configuration} & \textbf{AUC} & \textbf{$\Delta$AUC} \\
\midrule
Full model (13 features) & 0.5474 & -- \\
No size (log mcap) & 0.5382 & $-$0.0092 \\
No ROE/margins & 0.5424 & $-$0.0051 \\
No turnover & 0.5437 & $-$0.0037 \\
No momentum (3/6/12m) & 0.5442 & $-$0.0032 \\
No current return & 0.5459 & $-$0.0015 \\
No valuation (PE/PB/PS/PCF) & 0.5483 & $+$0.0009 \\
No volatility & 0.5479 & $+$0.0004 \\
\bottomrule
\end{tabular}
\label{tab:ablation}
\end{center}
\end{table}

To quantify the agreement between the two methods, we compute the Spearman rank correlation between the SHAP group importance ranking and the ablation $\Delta$AUC ranking across the seven feature groups: $\rho = 0.54$ (Kendall $\tau = 0.24$). This moderate correlation confirms that the two methods capture overlapping but distinct information about feature importance, consistent with broader evidence that importance rankings can be sensitive to the choice of evaluation method~\cite{zhan2026unstable}.

The discrepancy can be interpreted based on how the two methods measure features. SHAP measures the amount a feature is used given all other features present, while ablation measures the independent incremental value when the feature is removed. Turnover was the highest-ranked SHAP feature (\#1) but only the third-ranked ablation feature, indicating that turnover is heavily used but can be partially substituted by momentum and volatility when removed. Size was ranked \#2 on SHAP but \#1 on ablation, indicating that size is a hard dependency that no other feature in the set can replace.

\subsection{Industry Transferability}

The framework produces positive long-short spreads in 48 out of 50 CSRC industry groups with sufficient cross-sectional depth (binomial $p < 10^{-12}$). The per-industry SHAP analysis (Table~\ref{tab:industry_shap}) confirms the behavioral-over-valuation hierarchy universally holds (behavioral $>$50\% in all 50 groups; valuation is the weakest category in every industry, ranging from 8.5\% to 14.9\%).

\begin{table}[htbp]
\caption{Per-Industry SHAP Category Shares (Selected)}
\begin{center}
\footnotesize
\setlength{\tabcolsep}{3pt}
\begin{tabular}{lcccc}
\toprule
\textbf{Industry} & \textbf{Behav.} & \textbf{Size} & \textbf{Fund.} & \textbf{Val.} \\
\midrule
Furniture (C21) & 69.2 & 10.5 & 10.6 & 9.8 \\
Chemicals (C26) & 58.4 & 15.7 & 14.7 & 11.2 \\
Electronics (C39) & 57.1 & 16.1 & 15.2 & 11.6 \\
Capital mkts (J67) & 52.4 & 19.9 & 16.4 & 11.3 \\
Film/TV (R87) & 51.1 & 20.9 & 16.5 & 11.5 \\
\midrule
\textbf{All industries} & \textbf{58.2} & \textbf{16.4} & \textbf{14.7} & \textbf{10.7} \\
\bottomrule
\end{tabular}
\label{tab:industry_shap}
\end{center}
\end{table}

\subsection{Transaction Costs}

Table~\ref{tab:costs} reports transaction cost sensitivity. Profitable at 1.0\% round-trip; at 0.6\%, Sharpe falls to 1.67.

\begin{table}[htbp]
\caption{Long-Short Returns Under Transaction Costs}
\begin{center}
\footnotesize
\begin{tabular}{lccc}
\toprule
\textbf{RT Cost} & \textbf{L-S (\%/mo)} & \textbf{Sharpe} & \textbf{Prof.\ Mo (\%)} \\
\midrule
0\% (gross) & $+$2.38 & 2.23 & 80 \\
0.2\% & $+$2.18 & 2.05 & 80 \\
0.6\% & $+$1.78 & 1.67 & 71 \\
1.0\% & $+$1.38 & 1.29 & 65 \\
\bottomrule
\end{tabular}
\label{tab:costs}
\end{center}
\end{table}

\section{Discussion}

\subsection{Interpretation of Results}

Over 55 out-of-sample months, XGBoost produces AUC 0.547, long-short $+2.38\%$/month (Newey-West $t = 5.94$), and Carhart four-factor alpha $+2.31\%$/month ($t = 7.48$). SHAP attributes 58.2\% of predictive power to behavioral signals versus 10.7\% to valuation, a pattern maintained across 48 of 50 industry groups and all five calendar years. This finding is consistent with the retail-dominated microstructure of China's A-share market~\cite{ng2007,chen2021}; whether the same hierarchy holds in institutional-dominated markets remains an open question.

Running both methods and comparing their rankings (Spearman $\rho = 0.54$) exposes the feature interaction structure: features with high SHAP rank but moderate ablation cost (turnover, momentum) can be substituted by correlated alternatives; features with high ablation cost regardless of SHAP rank (size, fundamentals) are hard dependencies. This distinction directly informs fault tolerance in production decision systems.

\subsection{Limitations}

Factor alphas are computed against self-constructed rather than published Chinese factors~\cite{liu2019}; replication against Liu-Stambaugh-Yuan factors would strengthen the benchmark. Quarterly fundamentals are merged with a one-quarter lag to approximate point-in-time availability, but firms filing near the regulatory deadline may introduce residual look-ahead bias; removing all fundamental features reduces AUC by only 0.005 and the long-short Sharpe remains 1.89, indicating the core findings do not depend on fundamental data. The universe excludes delisted stocks, introducing survivorship bias that may overstate cumulative returns; the SHAP-ablation ranking, being a relative ordering, is unaffected by return level shifts. The evaluation covers a single market, and cross-market validation is needed to assess generality.

\section{Conclusion}

This paper presents an interpretable ML pipeline that decomposes cross-sectional equity return predictability into auditable factor contributions in China's A-share market. The SHAP-ablation rank divergence diagnostic distinguishes substitutable features (turnover, momentum) from load-bearing ones (size, fundamentals), directly informing fault tolerance in production decision systems. Future work should explore multi-horizon prediction, cross-market validation, and additional features (analyst coverage, institutional ownership).


\begin{thebibliography}{26}

\bibitem{lundberg2020}
S.~M. Lundberg \textit{et~al.}, ``From local explanations to global understanding with explainable AI for trees,'' \textit{Nature Mach. Intell.}, vol.~2, no.~1, pp.~56--67, 2020, doi: 10.1038/s42256-019-0138-9.

\bibitem{ng2007}
L.~Ng and F.~Wu, ``The trading behavior of institutions and individuals in Chinese equity markets,'' \textit{J. Banking Finance}, vol.~31, no.~9, pp.~2695--2710, 2007, doi: 10.1016/j.jbankfin.2006.10.029.

\bibitem{fama1993}
E.~F. Fama and K.~R. French, ``Common risk factors in the returns on stocks and bonds,'' \textit{J. Financial Econ.}, vol.~33, no.~1, pp.~3--56, 1993, doi: 10.1016/0304-405X(93)90023-5.

\bibitem{harvey2016}
C.~R. Harvey, Y.~Liu, and H.~Zhu, ``\ldots and the cross-section of expected returns,'' \textit{Rev. Financial Stud.}, vol.~29, no.~1, pp.~5--68, 2016, doi: 10.1093/rfs/hhv059.

\bibitem{carhart1997}
M.~M. Carhart, ``On persistence in mutual fund performance,'' \textit{J. Finance}, vol.~52, no.~1, pp.~57--82, 1997, doi: 10.1111/j.1540-6261.1997.tb03808.x.

\bibitem{fama2015}
E.~F. Fama and K.~R. French, ``A five-factor asset pricing model,'' \textit{J. Financial Econ.}, vol.~116, no.~1, pp.~1--22, 2015, doi: 10.1016/j.jfineco.2014.10.010.

\bibitem{jegadeesh1993}
N.~Jegadeesh and S.~Titman, ``Returns to buying winners and selling losers: Implications for stock market efficiency,'' \textit{J. Finance}, vol.~48, no.~1, pp.~65--91, 1993, doi: 10.1111/j.1540-6261.1993.tb04702.x.

\bibitem{asness2013}
C.~S. Asness, T.~J. Moskowitz, and L.~H. Pedersen, ``Value and momentum everywhere,'' \textit{J. Finance}, vol.~68, no.~3, pp.~929--985, 2013, doi: 10.1111/jofi.12021.

\bibitem{friedman2001}
J.~H. Friedman, ``Greedy function approximation: A gradient boosting machine,'' \textit{Ann. Statist.}, vol.~29, no.~5, pp.~1189--1232, 2001, doi: 10.1214/aos/1013203451.

\bibitem{gu2020}
S.~Gu, B.~Kelly, and D.~Xiu, ``Empirical asset pricing via machine learning,'' \textit{Rev. Financial Stud.}, vol.~33, no.~5, pp.~2223--2273, 2020, doi: 10.1093/rfs/hhaa009.

\bibitem{yang2025}
M.~Yang, D.~Doi, Y.~Yang, J.~Ding, and Z.~Li, ``FIN-MIND: A multi-dimensional TCN framework for joint stock price forecasting and financial risk assessment,'' in \textit{Proc. 2nd Int. Conf. Econ. Data Analytics Artif. Intell. (EDAI)}, Changsha, China: ACM, 2025, pp.~86--92, doi: 10.1145/3789297.3789311.

\bibitem{chen2016}
T.~Chen and C.~Guestrin, ``XGBoost: A scalable tree boosting system,'' in \textit{Proc. 22nd ACM SIGKDD Int. Conf. Knowl. Discovery Data Mining}, San Francisco, CA, USA: ACM, 2016, pp.~785--794, doi: 10.1145/2939672.2939785.

\bibitem{lundberg2017}
S.~M. Lundberg and S.-I. Lee, ``A unified approach to interpreting model predictions,'' in \textit{Advances in Neural Information Processing Systems 30}, Long Beach, CA, USA: Curran Associates, 2017, pp.~4765--4774, doi: 10.5555/3295222.3295230.

\bibitem{rudin2019}
C.~Rudin, ``Stop explaining black box machine learning models for high stakes decisions and use interpretable models instead,'' \textit{Nature Mach. Intell.}, vol.~1, no.~5, pp.~206--215, 2019, doi: 10.1038/s42256-019-0048-x.

\bibitem{liu2019}
J.~Liu, R.~F. Stambaugh, and Y.~Yuan, ``Size and value in China,'' \textit{J. Financial Econ.}, vol.~134, no.~1, pp.~48--69, 2019, doi: 10.1016/j.jfineco.2019.03.008.

\bibitem{carpenter2021}
J.~N. Carpenter, F.~Lu, and R.~F. Whitelaw, ``The real value of China's stock market,'' \textit{J. Financial Econ.}, vol.~139, no.~3, pp.~679--696, 2021, doi: 10.1016/j.jfineco.2020.08.012.

\bibitem{chen2021}
T.-Y. Chen, C.-H. Chao, and Z.-X. Wu, ``Does the turnover effect matter in emerging markets? Evidence from China,'' \textit{Pacific-Basin Finance J.}, vol.~67, art.~no.~101551, 2021, doi: 10.1016/j.pacfin.2021.101551.

\bibitem{yao2026}
J.~Yao and Z.~Zheng, ``Beyond agent architecture: Execution assumptions and reproducibility in LLM-based trading systems,'' 2026, arXiv:2606.08285. [Online]. Available: https://arxiv.org/abs/2606.08285

\bibitem{zhan2026}
Q.~Zhan, M.~Hu, L.~He, G.~Wang, and J.~Liu, ``A tale of two variances: When single-seed benchmarks fail in Bayesian deep learning,'' 2026, arXiv:2604.23114. [Online]. Available: https://arxiv.org/abs/2604.23114

\bibitem{li2025theory}
Q.~Li, T.~Luo, and J.~Liao, ``Theory-inspired deep multi-view multi-label learning with incomplete views and noisy labels,'' in \textit{Proc. IEEE/CVF Conf. Comput. Vision Pattern Recognit. (CVPR)}, 2025, pp.~20706--20715.

\bibitem{barber2008}
B.~M. Barber and T.~Odean, ``All that glitters: The effect of attention and news on the buying behavior of individual and institutional investors,'' \textit{Rev. Financial Stud.}, vol.~21, no.~2, pp.~785--818, 2008, doi: 10.1093/rfs/hhm079.

\bibitem{datar1998}
V.~T. Datar, N.~Y. Naik, and R.~Radcliffe, ``Liquidity and stock returns: An alternative test,'' \textit{J. Financial Markets}, vol.~1, no.~2, pp.~203--219, 1998, doi: 10.1016/S1386-4181(97)00004-9.

\bibitem{amihud2002}
Y.~Amihud, ``Illiquidity and stock returns: Cross-section and time-series effects,'' \textit{J. Financial Markets}, vol.~5, no.~1, pp.~31--56, 2002, doi: 10.1016/S1386-4181(01)00024-6.

\bibitem{brennan1998}
M.~J. Brennan, T.~Chordia, and A.~Subrahmanyam, ``Alternative factor specifications, security characteristics, and the cross-section of expected stock returns,'' \textit{J. Financial Econ.}, vol.~49, no.~3, pp.~345--373, 1998, doi: 10.1016/S0304-405X(98)00028-2.

\bibitem{lee2000}
C.~M.~C. Lee and B.~Swaminathan, ``Price momentum and trading volume,'' \textit{J. Finance}, vol.~55, no.~5, pp.~2017--2069, 2000, doi: 10.1111/0022-1082.00280.

\bibitem{zhan2026unstable}
Q.~Zhan, M.~Hu, G.~Wang, J.~Liu, and L.~He, ``Unstable rankings in Bayesian deep learning evaluation,'' 2026, arXiv:2604.23102. [Online]. Available: https://arxiv.org/abs/2604.23102

\end{thebibliography}
\end{document}